\documentclass[10pt,twocolumn,letterpaper]{article}

\usepackage[pagenumbers]{iccv} % To force page numbers, e.g. for an arXiv version
\usepackage{multirow}
\usepackage{placeins}
\usepackage{wrapfig}
\usepackage{tabularx}  % Add this in the preamble
\usepackage{stfloats}
\usepackage{afterpage}
\newcounter{desiredpage}
\usepackage{diagbox}
\definecolor{iccvblue}{rgb}{0.21,0.49,0.74}
\usepackage[pagebackref,breaklinks,colorlinks,allcolors=iccvblue]{hyperref}
\usepackage{placeins}
\usepackage{changes}
\usepackage{xcolor}
\definecolor{darkgreen}{rgb}{0.0, 0.5, 0.0}
\definechangesauthor[name={M}, color=blue]{M}
\definechangesauthor[name={N}, color=red]{N}
\definechangesauthor[name={R}, color=ForestGreen ]{R}
\definechangesauthor[name={A}, color=brown]{A}
\definechangesauthor[name={O}, color=purple]{O}

\title{Don't Judge Before You CLIP: \ \  A Unified Approach for Perceptual Tasks}

\author{
Amit Zalcher$^*$, Navve Wasserman$^*$, Roman Beliy, Oliver Heinimann, and Michal Irani \\
\normalsize Weizmann Institute of Science \\
\tt\small{*Indicates equal contribution.}
}

\begin{document}

\maketitle

\begin{abstract}

Visual perceptual tasks aim to predict human judgment of images (e.g., emotions invoked by images, image quality assessment). Unlike objective tasks such as object/scene recognition, perceptual tasks rely on {subjective} human assessments, making its data-labeling difficult. The scarcity of such human-annotated data  results in small datasets leading to poor generalization. Typically, specialized models were designed for each perceptual task, tailored to its unique characteristics and its own training dataset.  We propose a unified architectural framework for solving multiple different perceptual tasks leveraging CLIP as a prior. Our approach is based on recent cognitive findings which indicate that CLIP correlates well with human judgment. While CLIP was explicitly trained  to align images and text, it implicitly also learned human inclinations. We attribute this to the inclusion of human-written image captions in CLIP's training data, which contain not only factual image descriptions, but inevitably also human sentiments and emotions. This makes CLIP a particularly strong prior for perceptual tasks.  Accordingly, we suggest that minimal adaptation of CLIP  suffices for solving a variety of perceptual tasks. Our simple unified framework employs a lightweight adaptation to fine-tune CLIP to each task, without requiring any task-specific architectural changes. We evaluate our approach on three tasks: (i)~Image Memorability Prediction, (ii)~No-reference Image Quality Assessment, and (iii)~Visual Emotion Analysis. Our model achieves state-of-the-art results on all three tasks, while demonstrating improved generalization across different datasets. 
\end{abstract}

\section{Introduction}
\label{sec:intro}

\begin{figure*}[tp]
    \centering
    \includegraphics[width=\linewidth]{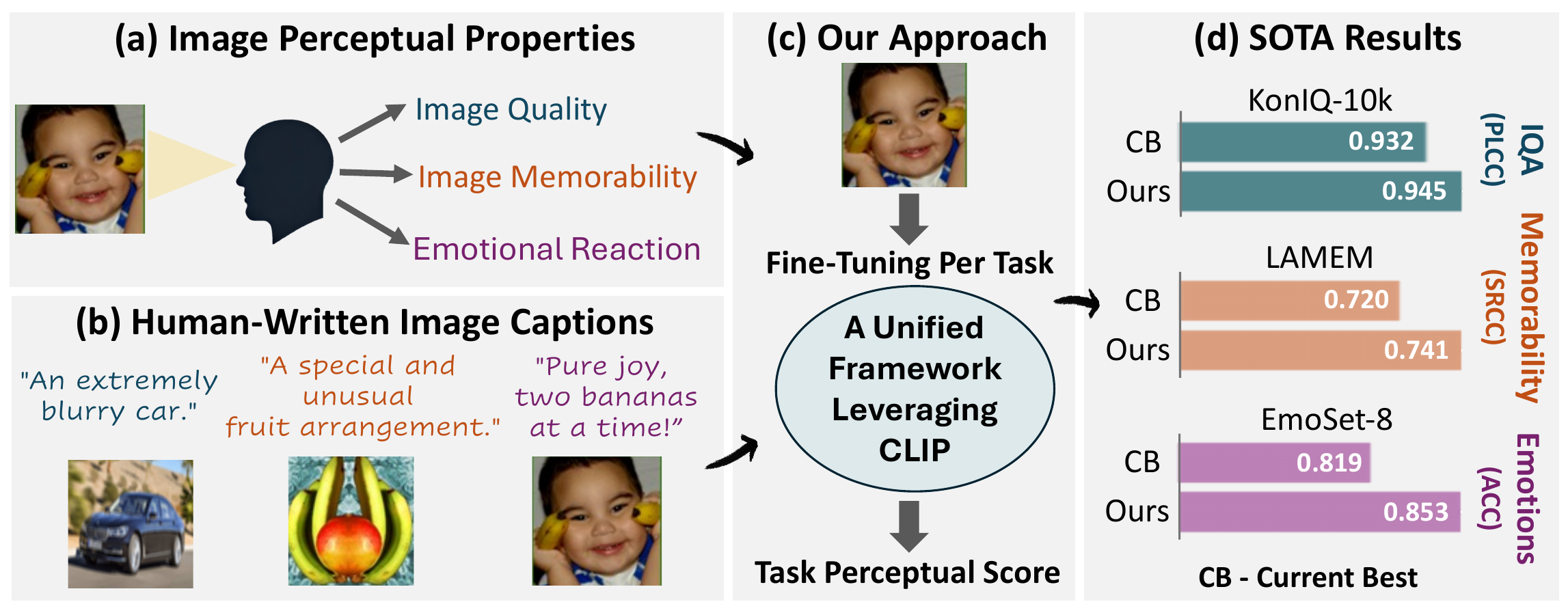}
    \vspace*{-0.4cm}
    \caption{\textbf{Our Framework:} (a)~Perceptual tasks rely on subjective human judgment. (b)~Illustration of CLIP’s training samples, which includes human-written captions. These human-generated annotations contain not only factual image descriptions, but inevitably also human sentiments, preferences and emotions. This suggests that CLIP can serve as a prior for perceptual tasks.
    (c)~Our approach leverages CLIP’s prior knowledge to address multiple perceptual tasks with minimal task-specific adaptation (d)~We achieve state-of-the-art performance across three distinct perceptual tasks. CB refers for the current best method in each task (see Tables \ref{tab:IQA_single},\ref{tab:MEM_single},\ref{tab:EMO_single} for numerical scores). 
    }
    \vspace{-0.17cm}
    \label{fig:teaser}
\end{figure*}

Visual perceptual tasks has been a long-standing research area \cite{marr2010vision,leeds2013comparing}, 
incorporating both psychological aspects \cite{gibson2014ecological,lu2011visual,shemyakina2024study} and practical applications like education~\cite{wang2021online,duivenboden2023effect} and advertising~\cite{pieters2010stopping,bi2024does,sample2020components,negm2015investigating}. These include predicting human-perceived image quality~\cite{khosla2015understanding,ma2017end,zhao2023quality}, emotional responses to images~\cite{peng2016emotions,yang2021solver}, and image memorability~\cite{isola2011makes,squalli2018deep}.
Unlike objective tasks such as object or scene recognition, perceptual tasks rely on \emph{subjective} human assessments, making its data-labeling difficult. First, capturing genuine human reactions requires carefully designed experiments, such as using precise survey questions to distinguish subtle emotional states evoked by images.
Additionally, variability in perception of different people necessitates collecting data from multiple participants to ensure robust and reliable results \cite{kret2012review,isola2011makes,fang2020perceptual}.
The scarcity of such human-annotated data results in small datasets which restricts the capabilities and generalization of predictive models.

\vspace{0.05cm}
Typically, specialized models were designed for each perceptual task, tailored to its unique characteristics and its own training dataset. For instance, emotion recognition models~\cite{xu2022mdan, xu2024learning} often integrate psychological insights, while image quality assessment models~\cite{xu2024boosting, qin2023data} focus on multi-level image features. However, fully understanding the specific characteristics of human perception for each perceptual task is an ongoing research and remains debatable~\cite{kramer2023features,chandler2013seven,zhang2024fmri,wang2023unlocking}. 

\vspace{0.05cm}
Since human-labeled data is limited, a strong \emph{perceptual} prior is needed. In this paper we propose a unified architectural framework, named {\mbox{\emph{\textbf{PerceptCLIP}}}}, for solving multiple different perceptual tasks.
Our approach is based on recent cognitive findings~\cite{shoham2024using} which indicate that CLIP correlates well with human judgment. 
While CLIP was explicitly trained to align images and text, it has also implicitly learned human preferences. We attribute this to the nature of CLIP’s training data, which includes human-written image captions that consist not only factual image descriptions, but inevitably also \emph{subjective} aspects such as human sentiments, preferences and emotions. This makes CLIP a particularly strong prior for \emph{perceptual tasks}. Accordingly, we suggest that \emph{minimal} adaptation of CLIP, rather than extensive fine-tuning, is sufficient to achieve strong performance on perceptual tasks. 

\vspace{0.05cm}
A few previous methods employed CLIP for specific perceptual tasks. However, some \emph{fully} fine-tuned most of its components~\cite{zhang2023blind}, thus erasing its trong pre-trained knowledge and risking dataset-overfitting. Others relied on tuning only CLIP's text-prompts ~\cite{deng2022simemotion,deng2024learning,wang2023exploring,xu2024learning}, which is too restrictive, preventing proper adaptation of relevant layers.  
\emph{Our approach strikes a balance between preserving CLIP’s strong ability to capture diverse perceptual attributes, while allowing sufficient flexibility for task-specific adaptation.} 

Specifically, our simple unified architectural framework employs a lightweight adaptation to fine-tune CLIP to each task, without requiring any task-specific architectural changes. 
We apply LoRA (Low-Rank Adaptation)~\cite{hu2021lora} on CLIP visual encoder, selectively 
\emph{fine-tuning only the attention layers}, followed by an MLP head. This approach preserves CLIP’s perceptual priors while enabling task-specific refinement.

\vspace{0.04cm}
We evaluate our approach on three tasks: (i) Image Memorability Prediction, (ii) No-reference Image Quality Assessment, and (iii) Visual Emotion Analysis. Our model achieves state-of-the-art results on all three tasks, while demonstrating improved generalization across different datasets.
Our results show that CLIP already holds rich perceptual knowledge and, with efficient tuning, surpasses previous task-specific methods without extensive manual adjustments or domain expertise. 
Furthermore, we demonstrate the benefits of joint training on multiple datasets of the same task, by using a different MLP head per dataset, with a joint finetuned CLIP backbone  for all datasets. This significantly enhances model performance on small datasets.

\vspace*{0.2cm}
\noindent\textbf{Our contributions are as follows:}
\begin{itemize}
    \item We propose a \textit{simple unified framework} that leverages CLIP’s pre-trained knowledge, enabling effective adaptation to multiple perceptual tasks and \textit{eliminating the need for task-specific architectural adjustments.} 
    \item Our models achieve \textit{state-of-the-art performance across 3  different perceptual tasks}: image memorability prediction, no-reference image quality assessment, and visual emotion analysis, showcasing both superior results, as well as better generalization across different datasets.
    \item We demonstrate that joint training on multiple different datasets of the same task, leads to enhanced model performance on small datasets. 
\end{itemize}

\section{Related Work}
\label{sec:related}

\paragraph{No-Reference Image Quality Assessment (IQA).} 
No-reference IQA involves predicting the perceptual quality of an image (according to human judgment) without a reference image. Early methods predominantly used CNN-based architectures~\cite{kang2014convolutional,ma2017end,su2020blindly,networkblind,saha2023re} or hybrid CNN-transformer models~\cite{you2021transformer,xu2024boosting}. These approaches aimed to capture both local and global image features, essential for accurate quality assessment~\cite{qin2023data}. More recent transformer-based models have demonstrated further improvement~\cite{qin2023data,ke2021musiq}.  
A notable advancement in IQA has been the integration of large-scale pretrained models~\cite{zhang2023blind,xu2024boosting,wang2023exploring}. While CLIP has been utilized as a pretrained backbone~\cite{zhang2023blind,wang2023exploring}, existing approaches either fully fine-tune it or optimize only text prompts, failing to effectively adapt CLIP to the task and fully leverage its strong prior. The current state-of-the-art~\cite{xu2024boosting} uses both a pretrained ViT and ResNet trained on ImageNet, adapting them to the IQA task. 
Our method surpasses these approaches while having significantly fewer trainable parameters.

\paragraph{Image Memorability Prediction.} 
Early works~\cite{isola2011understanding,isola2011makes} demonstrated that image memorability (i.e., the likelihood of an image to be remembered) is an inherent property of an image, influenced primarily by its content and structure. Those initial approaches relied on global image descriptors and color histograms, later evolving to leverage features extracted from fine-tuned CNNs~\cite{khosla2015understanding}, which significantly improved prediction performance. Subsequent methods~\cite{squalli2018deep,leonardi2019image} incorporated additional semantic features from image captioning systems or soft attention mechanisms. Residual networks (ResNets) further enhanced memorability estimation, with studies like~\cite{fajtl2018amnet} integrating ResNet50 and LSTM for regression. 
A recent work~\cite{hagen2023image} explored using Vision Transformer (ViT)~\cite{dosovitskiy2020image}, which performed well but still fell short of ~\cite{squalli2018deep}.
Our approach achieves superior performance compared to the above methods leveraging CLIP perceptual prior.

\paragraph{Visual Emotion Analysis.} 
The problem of assessing the emotional reaction and sentiment of people towards images has evolved through various approaches. Early works~\cite{yang2018weakly,rao2019multi} used pre-trained CNN backbones for feature extraction, while graph-based methods~\cite{yang2021solver,wu2021discovering,yang2021stimuli} emphasized object-graph relationships and emotion-enhanced features, highlighting the importance of contextual interactions. \cite{xu2022mdan} introduced hierarchical emotion modeling, leveraging psychological theories for improved emotion representation. CLIP has recently been explored for emotion recognition~\cite{deng2022simemotion,deng2024learning,xu2024learning}, mainly through prompt-based learning while some further use insights from psychological research.
Our method demonstrates improved results, without relying on task-specific knowledge. \\

\vspace{-0.1cm}
\noindent
Unlike previous methods that rely on task-specific architectures, we use a single architecture across tasks, eliminating the need for extensive modifications or domain expertise. 
As described above, few works use CLIP for specific perceptual tasks. However, they either trained all of it's parameters, risked the loss of its priors or used a restrictive approach by applying only prompt tuning. In contrast, our approach strikes a balance between preserving CLIP's perceptual prior and allowing task-specific adaptation, achieving state-of-the-art results.

\section{A Unified Framework for Perceptual Tasks}
\label{sec:method}

\begin{figure}[tp]
    \centering
    \includegraphics[width=\linewidth]{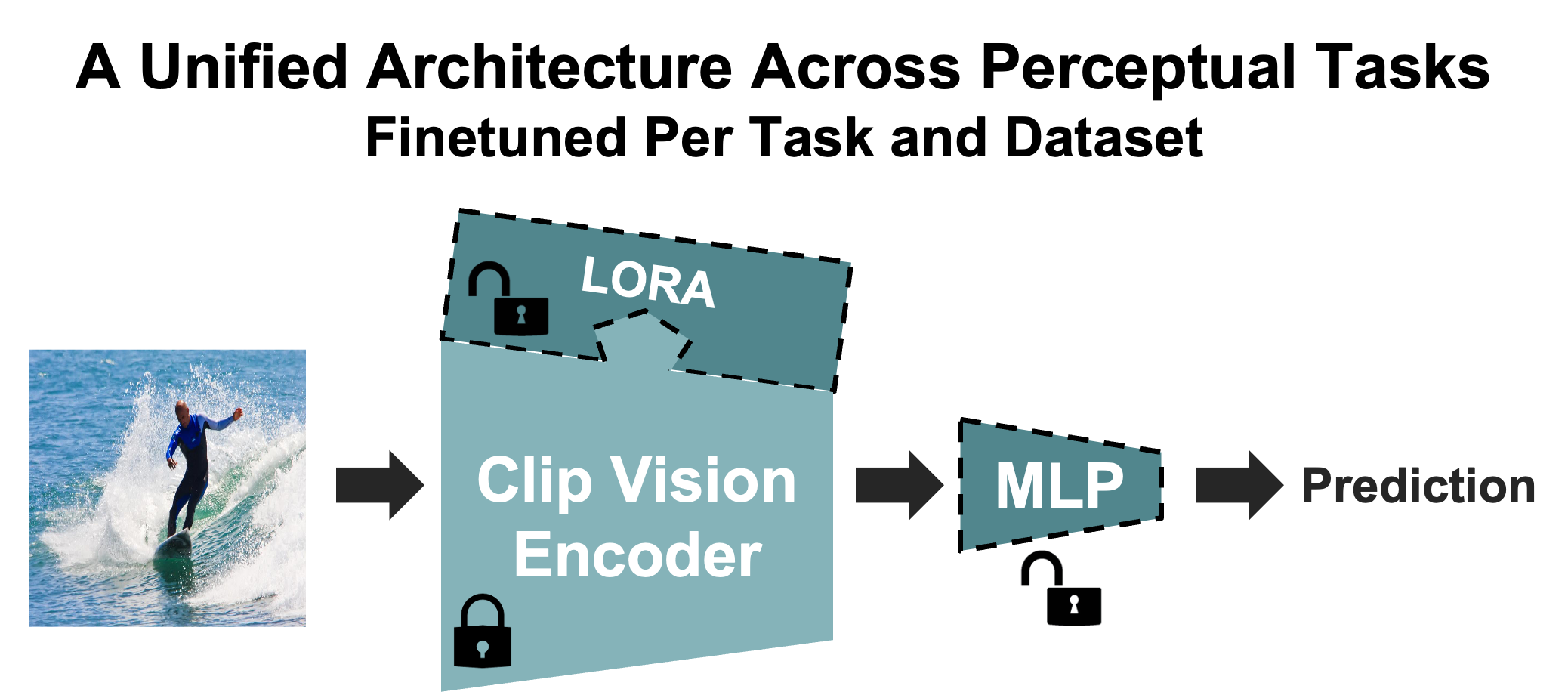}
\caption{\textbf{Unified Framework for Perceptual Tasks:} Leveraging the CLIP vision encoder, following an MLP head, our architecture maintains a simple, shared structure across diverse perceptual tasks. With lightweight LoRA adaptation, it fine-tunes efficiently for each task independently, effectively exploiting CLIP’s prior perceptual knowledge.}
\vspace{-0.1cm}
\label{fig:unified_framework}
\end{figure}

We propose a \emph{unified framework} that \emph{adapts effectively to diverse perceptual tasks} while maintaining a simple, consistent architecture (see \cref{fig:unified_framework}). Our approach uses the CLIP vision encoder~\cite{radford2021learning}, followed by an MLP head for task-specific predictions. To preserve CLIP’s pretrained knowledge, we apply LoRA~\cite{hu2021lora} to the attention weights, enabling lightweight task-specific adaptation with minimal additional parameters. We use the same architecture and set of hyperparameters across all tasks, reducing the need for extensive task-specific adjustments. In \cref{sec:experiments}, we evaluate our approach on three distinct perceptual tasks: Visual Emotion Analysis, Memorability Prediction, and Image Quality Assessment.

\subsection{Architecture and Training}
To enable efficient adaptation while preserving CLIP’s perceptual knowledge, we fine-tune only the query (q), key (k), and value (v) attention layers of the vision transformer using LoRA, with 
r=16 (decomposition rank) and $\alpha$=8 (scaling factor).
This significantly reduces the number of trainable parameters while allowing task-specific adaptations, resulting in \textit{less than} 3M trainable parameters.

\paragraph{Model Training}
 We follow a consistent training strategy across all tasks. The LoRA weights and the MLP are optimized using the AdamW optimizer with a weight decay of 1e-4. We explore 4 different learning rates (5e-5, 1e-4, 5e-4, 1e-3) and implement early stopping with a patience of 12 epochs and maximum of 40 epochs. The final model is selected based on the highest validation score. Before being input into the model, all images are normalized using the CLIP preprocessor. Task specific losses and augmentations are explained in \cref{sec:tasks} and Supplementary \ref{sup:Technical}.

\subsection{Multi Dataset Training}
\label{subsec:multi_dataset_approach}
While our method performs well, generalizing from training to testing can still be challenging with small datasets.
We propose adapting the same CLIP vision encoder using LoRA across multiple datasets within the same task, allowing the model to leverage larger datasets and improve performance on smaller ones (see \cref{sec:experiments}).

Our approach uses a single CLIP vision encoder with LoRA weights, along with dataset-specific MLP heads. The training procedure follows a two-stage process. In the first stage, the entire model is trained together, with proportional sampling ensuring that each dataset contributes to a batch in proportion to its size. The model is then optimized by averaging dataset-specific losses. In the second stage, only the MLP heads are fine-tuned towards the target dataset while the CLIP encoder remains frozen. This allows our model to adapt to each dataset with its unique characteristics without altering the shared representations provided by the CLIP encoder (which is shared by all datasets). 

\section{Tasks And Datasets}
\label{sec:tasks}
\begin{figure}[tp]
    \centering
    \includegraphics[width=1.\linewidth]{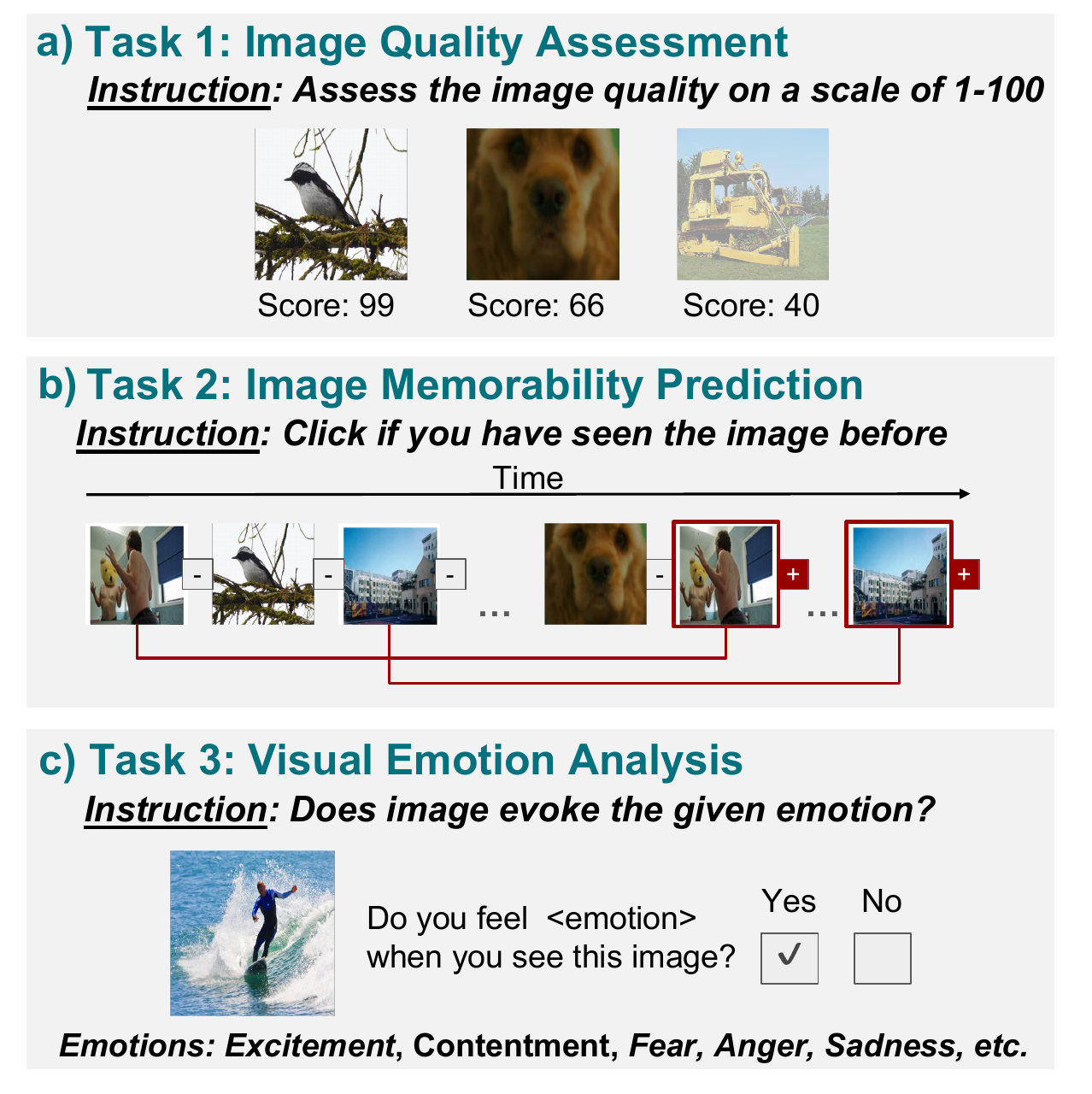}
    \vspace{-0.6cm}
    \caption{\textbf{Visual Perceptual Tasks.}}
    \label{fig:tasks}
    \vspace{-0.2cm}
\end{figure}

In this section, we present the three perceptual tasks which were used in our work (\cref{fig:tasks}), along with their dataset details and loss functions (with further details in Supplementary \ref{sup:Technical}).

\subsection{Image Quality Assessment \ \emph{(without a Reference)}}
Image Quality Assessment (IQA) measures how humans perceive image quality, considering factors like sharpness, noise, and distortions, and is important for many visual tasks (e.g. image generation and enhancement). It is formulated as a regression problem, predicting a continuous quality score that aligns with human judgment (see Fig.~\ref{fig:tasks}a).
We use diverse benchmark datasets with both authentic distortions (e.g., motion blur, overexposure) and synthetic distortions  (e.g., manually added noise and color quantization) of varying types and intensities. Each dataset is annotated via crowd-sourced quality ratings, with mean opinion scores (MOS) reflecting human perception. 

We used four \emph{\textbf{authentic distortion}} benchmark datasets: LIVEC~\cite{ghadiyaram2015massive} (1,162 images) and SPAQ~\cite{fang2020perceptual} (11,125 images), both focused on real images captured by mobile devices; KonIQ-10k~\cite{hosu2020koniq} (10,073 images), which includes diverse real-world scenes; and FLIVE~\cite{ying2020patches}, comprising 39,810 images from social media and streaming platforms. 
We further used three \emph{\textbf{synthetic distortions}} benchmark datasets, which include distortions such as compression artifacts, noise, and color quantization. Each dataset consists of reference images and corresponding controlled degradations: TID2013~\cite{ponomarenko2013color} contains 25 reference images and 3,000 distorted images, LIVE~\cite{sheikh2006statistical} includes 29 reference images and 779 distorted images, and KADID-10k~\cite{lin2019kadid} provides 81 reference images and 10,125 distorted images.

We follow the same training procedure as \cite{xu2024boosting}, including the loss function, augmentations, and data splits (see Supplementary \ref{sup:Technical}). Specifically, we use a loss function based on the Pearson Linear Correlation Coefficient (PLCC). Given a batch of m images, with predicted quality scores $\tilde{y} = {\tilde{y}_1, \tilde{y}_2, ..., \tilde{y}_m}$ and corresponding ground-truth labels $y = {y_1, y_2, ..., y_m}$, the PLCC-induced loss is defined as: 
$L_{\text{PLCC}} = \frac{1}{2} (1 - \text{PLCC}(\tilde{y}, y))$
where \(\text{PLCC}(\tilde{y}, y)\) represents the Pearson correlation coefficient between the predicted and ground-truth scores.
During inference, we average the score predictions from 15 randomly cropped versions of each image. The data is split 10 times, with each split using 80\% for training and 20\% for testing. From the training set, 10\% is further reserved for validation.

\vspace{0.4cm}
\begin{table*}[t!]
\resizebox{\textwidth}{!}{%
\begin{tabular}{lcccccccc||cccccccccccc}
    \toprule
    \multirow{2}{*}{\textbf{Method}} & \multicolumn{2}{c}{\textbf{LIVEC}} & \multicolumn{2}{c}{\textbf{KonIQ-10k}} & \multicolumn{2}{c}{\textbf{SPAQ}} & \multicolumn{2}{c||}{\textbf{FLIVE}} & \multicolumn{2}{c}{\textbf{LIVE}} & \multicolumn{2}{c}{\textbf{TID2013}} & \multicolumn{2}{c}{\textbf{KADID-10k}} \\
    \cmidrule(r){2-3} \cmidrule(r){4-5} \cmidrule(r){6-7} \cmidrule(r){8-9} \cmidrule(r){10-11} \cmidrule(r){12-13} \cmidrule(r){14-15}
    & {SRCC} & {PLCC} & {SRCC} & {PLCC} & {SRCC} & {PLCC} & {SRCC} & {PLCC} & {SRCC} & {PLCC} & {SRCC} & {PLCC} & {SRCC} & {PLCC} \\
    \midrule
    HyperIQA (27M) & 0.852 & 0.882 & 0.906 & 0.917 & 0.911 & 0.915 & 0.544 & 0.602 & 0.962 & 0.966 & 0.840 & 0.858 & 0.852 & 0.845 \\
    MUSIQ (27M) & 0.702 & 0.746 & 0.916 & 0.928 & 0.918 & 0.921 & 0.566 & 0.661 & 0.940 & 0.911 & 0.773 & 0.815 & 0.875 & 0.872 \\
    TREs (152M) & 0.846 & 0.877 & 0.915 & 0.928 & - & - & 0.544 & 0.625 & 0.969 & 0.968 & 0.863 & 0.883 & 0.859 & 0.858 \\
    DEIQT (24M) & 0.875 & 0.894 & 0.921 & 0.934 & 0.919 & 0.923 & 0.571 & 0.663 & \underline{\textbf{0.980}} & \underline{\textbf{0.982}} & \underline{\textbf{0.892}} & \underline{\textbf{0.908}} & 0.889 & 0.887 \\
    LIQE (151M) & \textbf{0.904} & 0.910 & 0.919 & 0.908 & - & - & - & - & 0.970 & 0.951 & - & - & 0.930 & 0.931 \\
    Re-IQA (24M) & 0.840 & 0.854 & 0.914 & 0.923 & 0.918 & 0.925 & - & - & 0.970 & 0.971 & 0.804 & 0.861 & 0.872 & 0.885 \\
    QPT-ResNet50 (24M) & 0.895 & \textbf{0.914} & 0.927 & 0.941 & 0.925 & 0.928 & 0.575 & 0.675 & - & - & - & - & - & - \\
    LoDa (9M) & 0.876 & 0.899 & \textbf{0.932} & \textbf{0.944} & \textbf{0.925} & \textbf{0.928} & \textbf{0.578} & \textbf{0.679} & 0.975 & 0.979 & 0.869 & 0.901 & \textbf{0.931} & \textbf{0.936} \\
    \midrule
    \textbf{PerceptCLIP (Ours) (3M)} & \underline{\textbf{0.908}} & \underline{\textbf{0.923}} & \underline{\textbf{0.945}} & \underline{\textbf{0.954}} & \underline{\textbf{0.930}} & \underline{\textbf{0.933}} & \underline{\textbf{0.590}} & \underline{\textbf{0.689}} & \underline{\textbf{0.980}} & \underline{\textbf{0.982}} &\textbf{0.887} & \underline{\textbf{0.908}} & \underline{\textbf{0.952}} & \underline{\textbf{0.956}} \\
    \bottomrule
\end{tabular}}
\vspace{0.1cm}
\caption{\textbf{Model Performance on IQA Benchmarks.} Models are evaluated on seven common Image Quality Assesment benchmarks, reporting the median SRCC (Spearman's Rank Correlation) and PLCC (Pearson Linear Correlation) across 10 splits, along with the number of trainable parameters. The table shows that our model, PerceptCLIP, outperforms all leading IQA-dedicated methods, achieving the best performance, on six out of seven datasets while using significantly fewer trainable parameters.}
\vspace{-0.05cm}
\label{tab:IQA_single}
\end{table*}

\subsection{Image Memorability Prediction} \label{sec:methods_memorability}

This task aims to predict how well an image is remembered. It is formulated as a regression problem, where the goal is to estimate a memorability score reflecting the likelihood of the image being remembered (see Fig.~\ref{fig:tasks}b).

We train and evaluate our model on two benchmark datasets. The LaMem dataset~\cite{khosla2015understanding} contains a large-scale collection of $\sim$60,000 
images from diverse sources, including scenes, objects, and art. The Things memorability dataset~\cite{kramer2023features} comprises of 26,107 object images covering 1,854 distinct concepts. 
The LaMem dataset is split into five parts: 75\% for training, 5\% for validation, and 20\% for testing. For the Things dataset, we use the entire dataset for generalization assessment. 
The loss used for training is the mean squared error (MSE) loss on the memorability score.

\subsection{Visual Emotion Analysis} \label{sec:methods_emotions}
Visual Emotion Analysis (VEA) is a classification task focused on predicting the emotional response that an image is likely to evoke in viewers (see Fig.~\ref{fig:tasks}c).

We experiment with two prominent benchmark datasets, both supporting binary emotion classification (positive vs. negative) and multi-class classification of more nuanced emotions, carefully labeled by human annotators. EmoSet~\cite{yang2023emoset}, the largest existing visual emotion dataset, contains 118,102 images from social networks and artistic sources, labeled into 8 balanced emotions: Awe, Excitement, Amusement, and Contentment (positive) / Anger, Sadness, Disgust, and Fear (negative). EmotionROI~\cite{peng2016emotions} includes 1,980 Flickr annotated images, categorized into 6 emotions: Joy, Surprise (positive) / Anger, Disgust, Fear, and Sadness (negative). Following~\cite{yang2023emoset}, EmoSet is split into 80\% training, 15\% test, and 5\% validation, while EmotionROI is randomly divided into 5 splits with 75\% training, 20\% test, and 5\% validation. Our model is trained using the cross-entropy loss.

\section{Experimental Results}
\label{sec:experiments}

In this section, we present the results of our proposed approach for training models on three different perceptual tasks. Our trained models achieved state-of-the-art (SOTA) results with significant improvements across all three tasks. We first present the results for each task, comparing them against the best task-specific methods. We then demonstrate the effectiveness of training a model with multiple datasets for a given task, leading to substantial performance gains on smaller datasets.

\subsection{Image Quality Assessment}
\vspace{0.1cm}
We train and evaluate our model on 7 prominent datasets, covering both authentic and synthetic distortions, using the standard 10 splits provided by \citet{xu2024boosting}. Our approach is evaluated using the median Spearman’s Rank Correlation Coefficient (SRCC) and median Pearson’s Linear Correlation Coefficient (PLCC) with a 4-parameter logistic regression correction, as in \cite{xu2024boosting} (see Supplementary \ref{4_parameter_correction}).

\Cref{tab:IQA_single} summarizes our model’s performance in predicting image quality scores compared to leading IQA-dedicated methods \cite{su2020blindly,ke2021musiq,kramer2023features,golestaneh2022no,qin2023data,zhang2023blind,saha2023re,zhao2023quality,xu2024boosting}, along with the number of trainable parameters for each model. Our model achieves the best results on 6 out of 7 datasets in both SRCC and PLCC. On the remaining dataset (TID2013), it achieves a comparable PLCC score but slightly 
lower SRCC compared to the top-performing IQA-dedicated model (DEIQT~\cite{qin2023data}). Our results are even more impressive when considering the number of trainable parameters. Our model consist approximately 3M trainable parameters, compared to 9M to 152M in other models, highlighting the strength of CLIP’s perceptual prior.

\vspace{0.03cm}
Furthermore, as shown in \Cref{tab:IQA_gen}, our method not only sets new state-of-the-art results but also demonstrates strong generalization capabilities \emph{across different datasets} (i.e., training on one dataset, but testing on another).
This highlights the robustness of our model and its ability to learn meaningful perceptual features, rather than just overfitting dataset-specific patterns.
Notably, the improvement in generalization is significantly larger than the improvement in within-dataset evaluation. For instance, we achieve a 1.4\% relative SRCC improvement compared to LoDA when training and testing on KonIQ, while in a generalization setting, when training on FLIVE and testing on KonIQ, the improvement increases to 6.4\% relative SRCC. Our lightweight adaptation ensures that we retain CLIP's strong human-like perceptual priors when fine-tuning for specific tasks.

\begin{table}[h!]
\centering
\resizebox{0.47\textwidth}{!}{%
\begin{tabular}{llcccc}
\toprule
\textbf{Train dataset} &  \multicolumn{2}{c}{\textbf{FLIVE}}& \textbf{LIVEC} & \textbf{KonIQ}\\ 
        \midrule
\textbf{Test dataset} & \textbf{KonIQ} &  \textbf{LIVEC} & \textbf{KonIQ} &  \textbf{LIVEC} \\ 
\midrule
HyperIQA &     0.758&0.735& 0.772&0.785\\ 
TReS   &        0.713&0.740&0.733&0.786\\ 
DEIQT  &        0.733&0.781&0.744&0.794\\
QPT-ResNet50   &-&-&0.749&0.821\\ 
LoDa   &        0.763&0.805&0.745&0.811\\

\midrule
\textbf{PerceptCLIP (Ours)}&\textbf{0.812}&\textbf{0.825}&\textbf{0.794}&\textbf{0.875} \\
\bottomrule

\end{tabular}
}
\caption{\textbf{Cross-Dataset Generalization in IQA.} SRCC generalization results for training on one dataset and testing on another. PerceptCLIP significantly outperforms all previous models.
}

\label{tab:IQA_gen}
\end{table}
\begin{table*}[th!]
    \hspace{-0.35cm}
    \centering
    \begin{minipage}{0.5\textwidth}
        \centering
        {\small
        \begin{tabular}{lcc}
            \toprule
            \multirow{2}{*}{\textbf{Method}} 
            & \multicolumn{2}{c}{\textbf{LaMem}} \\
            \cmidrule(r){2-3} 
            & {SRCC$\uparrow$} & {MSE$\downarrow$} \\
            \midrule
            MemNet  
            & 0.640 & Unknown \\
            AMNet  
            & 0.677 & 0.0082  \\
            Leonardi et al  
            & 0.687 & 0.0079  
             \\ 
            ViTMem  
            & 0.711 & 0.0076  
              \\ 
            Squalli-Houssaini et al
            & 0.720 & 0.0092  
            \\ 
            \midrule
            \textbf{PerceptCLIP (Ours)}  
            & \textbf{0.741} & \textbf{0.0069} \\ 
            \bottomrule
        \end{tabular}
        }
        \parbox{0.9\textwidth}
        {\vspace{0.35cm} \caption{\textbf{Model Performance on Image Memorability.} Average SRCC and MSE across 5 splits of the LAMEM dataset are reported, 
        showing that our model, PerceptCLIP, surpasses all previous methods.}
        \label{tab:MEM_single}}
    \end{minipage}
    \hfill
    \begin{minipage}{0.48\textwidth}
        \centering
        \resizebox{1\textwidth}{!}{%
        \begin{tabular}{lcccc}
            \toprule
            \multirow{2}{*}{\textbf{Method}} 
            & \multicolumn{2}{c}{\textbf{EmoSet}} 
            & \multicolumn{2}{c}{\textbf{EmotionROI}} \\ 
            \cmidrule(r){2-3} 
            \cmidrule(r){4-5}  
            & \textbf{8} & \textbf{2} 
            & \textbf{6} & \textbf{2}  \\ 
            \midrule
            MRCNN  
            & 0.7539 & 0.9228  
            & - & -  \\ 
            MDAN  
            & 0.7575 & 0.9371  
            & 0.6166 & -  \\ 
            StimuliVEA  
            & 0.7840 & 0.9458  
            & 0.6162 & -  \\ 
            PT-DPC  
            & 0.7713 & 0.9246  
            & 0.6970 & 0.8855  \\ 
            SimEmotion  
            & 0.7906 & 0.9428  
            & 0.7054 & 0.9040  \\ 
            MVP
            & 0.8192 & 0.9644  
            & 0.7189 & 0.9255  \\ 
            \midrule
            \textbf{PerceptCLIP (Ours)}  
            & \textbf{0.8528} & \textbf{0.9780}  
            & \textbf{0.7485} & \textbf{0.9288}  \\ 
            \bottomrule
        \end{tabular}
        }
        \parbox{1\textwidth}{\vspace{0.35cm} \caption{\textbf{Model Performance on Emotion Classification.} Accuracy results for binary and multi-class emotion classification on the EmotionROI (mean over five splits) and EmoSet datasets. 
        Our model achieves state-of-the-art performance for both datasets.
        } 
        \label{tab:EMO_single}}
    \end{minipage}
    \vspace{-0.5cm}
\end{table*}

\vspace{-0.1cm}
\begingroup
\begin{table*}[b]
    \centering
    \resizebox{0.9\textwidth}{!}{%
    \begin{tabular}{lccccccc}
        \toprule
        \multirow{3}{*}{\textbf{Method}}
        & \multicolumn{4}{c}{\textbf{IQA}}
        & \textbf{VEA}
        & \multicolumn{2}{c}{\textbf{Memorability prediction}}
        \\
            \cmidrule(r){2-5} 
        \cmidrule(r){6-6}
            \cmidrule(r){7-8}

        & \multicolumn{2}{c}{\textbf{LIVEC}}
        & \multicolumn{2}{c}{\textbf{TID2013}}
        & \textbf{EmoROI-6} 
        & \multicolumn{2}{c}{\textbf{THINGS}} \\ 
        \cmidrule(r){2-3} 
        \cmidrule(r){4-5} 
        \cmidrule(r){6-6}
        \cmidrule(r){7-8} 
        & {SRCC$\uparrow$} & {PLCC$\uparrow$}
        & {SRCC$\uparrow$} & {PLCC$\uparrow$} 
        & {ACC$\uparrow$}  
        & {SRCC$\uparrow$} & {MSE$\downarrow$}  \\ 
        \midrule
        \textbf{Previous SOTA} 
        & 0.904 & 0.910 
        & 0.892 & 0.908
        & 0.7189  
        & - & - 
         \\
        \textbf{PerceptCLIP (Ours)} 
        & 0.908 & {0.923} 
        & 0.887 & 0.908
        & {0.7485}  
        & {0.452} & {0.0058} 
         \\
    \textbf{PerceptCLIP \small Multi-Dataset \normalsize (Ours)}
    
        & \textbf{0.922} & \textbf{0.933} 
        & \textbf{0.900} & \textbf{0.915} 
    
        & \textbf{0.7591} 
        & \textbf{0.454} &  \textbf{0.0054} 
        \\
    
        \bottomrule
    \end{tabular}
    }
    \parbox{0.9\textwidth}{\vspace{0.2cm}\caption{\textbf{Multi-Dataset Training Improves Performance.} The table shows the benefits of training on multiple datasets within the same task, demonstrating significantly improved performance on smaller datasets. Results for all datasets (shown in the Supplementary) show that these multi-dataset models achieve state-of-the-art results across all benchmarks.}
    \label{tab:multi_datasets}}
\end{table*}
\endgroup
% }

\subsection{Image Memorability Prediction} 
\vspace{-0.03cm}
We train and evaluate our model on the LaMem dataset using the standard 5-fold validation strategy. To assess generalization, we evaluate our model on the THINGS memorability dataset, as done in ~\cite{needell2022embracing,hagen2023image}. Performance is reported using the mean SRCC and averaged mean squared error (MSE) across 5 folds.  
\Cref{tab:MEM_single} presents our model’s performance on LaMem compared to leading memorability-dedicated methods~\cite{khosla2015understanding,fajtl2018amnet,leonardi2019image,hagen2023image,squalli2018deep}. Our model significantly surpasses all others, achieving new state-of-the-art results on both metrics. Specifically, it improves SRCC by 2.9\% relative to the second-best model~\cite{squalli2018deep} and reduces MSE by 9.2\% relative to the previous best-performing model~\cite{hagen2023image}.  
Furthermore, our model demonstrates significantly better generalization capabilities. When trained on LaMem and evaluated on THINGS, it achieves a 12\% relative improvement over the best previously reported result~\cite{hagen2023image}, as shown in the supplementary \Cref{sup:MEM_gen}. Altogether, this demonstrates the power of CLIP as a prior for the image memorability task.

\subsection{Visual Emotion Prediction}  
We train and evaluate on EmoSet and EmotionROI datasets, using the standard splits as described in \cref{sec:methods_emotions}, and report emotion classification accuracy for 2, 6, and 8 emotion categories, depending on the dataset.  
\Cref{tab:EMO_single} compares our model’s performance with various existing methods~\cite{rao2019multi,xu2022mdan,yang2021stimuli,deng2022simemotion,deng2024learning,xu2024learning} designed to classify the emotions evoked by images. The results demonstrate that our model achieves state-of-the-art performance on both benchmarks, excelling in both binary emotion classification (positive vs. negative) and multi-class emotion classification (8 categories for EmoSet, and 6 for EmotionROI).  

Notably, the improvement in multi-class emotion classification is more significant, as binary classification already exhibits high performance across models. Our model achieves an approximately 4\% relative accuracy improvement for both EmoSet and EmotionROI in multi-class emotion classification. These results further strengthen the effectiveness of our approach. Additional comparisons and metrics can be found in Supplementary \ref{sup:results}, in tables \ref{tab:EMO_metrics_1} and \ref{tab:EMO_metrics_2}.

\begin{figure*}[tp]
    \vspace{0.3cm}
    \centering
    \includegraphics[width=1\linewidth]{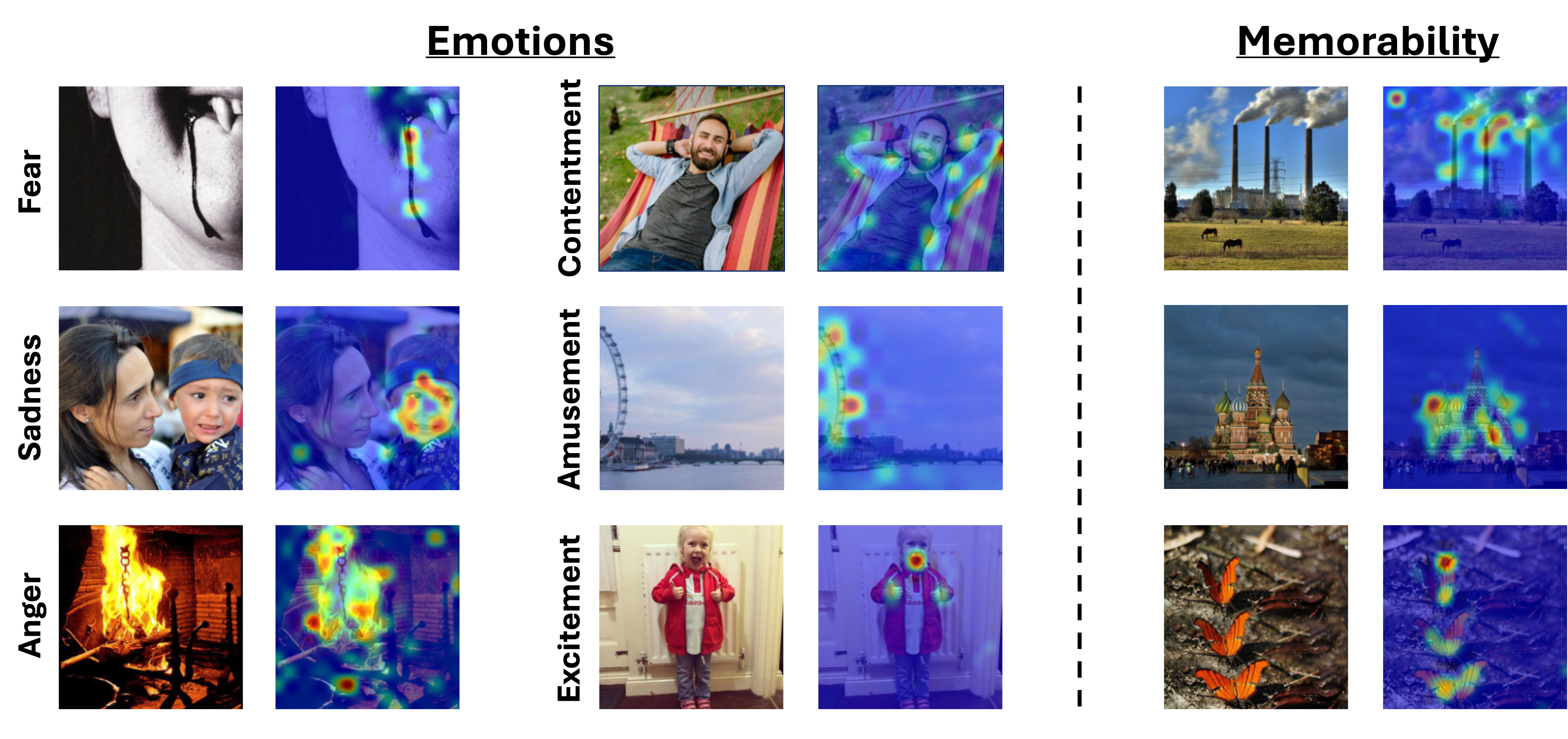}
    \vspace*{-0.4cm}
    \caption{\textbf{Attention Shift Toward Perceptual Cues.} 
    We present images along with the differences in their attention maps between our PerceptCLIP model and the pretrained CLIP vision encoder (displaying results from critical attention heads that most influence the perceptual predictions). This highlights the shift in attention, revealing how our model reallocates focus to perceptually meaningful regions.}
    \label{fig:visulaization}
\end{figure*}     

% ֿ\newpage
\vspace{0.1cm}
\subsection{Multi-Dataset Training}
\label{subsec:multi_datasets_results}
We present the results of our multi-dataset training, where a single model is trained on multiple datasets of the same task. In this setup, the LoRA finetuning of CLIP's attention layers is shared across datasets, while the final added MLP heads are dataset-specific.

Training on multiple datasets provides additional data, significantly improving performance on smaller datasets (see \Cref{tab:multi_datasets}) while maintaining comparable results on larger datasets (see supplementary \Cref{tab:Multi_EMO,tab:Multi_MEM,tab:Multi_IQA}). 
We present the results of four models, each trained on a specific set of datasets: (1) authentic IQA datasets (LIVEC, KonIQ-10k, SPAQ), (2) synthetic IQA datasets (KADID-10K, LIVE, TID-2013), (3) memorability datasets (THINGS, LaMem), and (4) emotion datasets (EmotionROI, EmoSet).
For example, in the IQA task we achieve a 1.5\% relative improvement on the small LIVEC dataset (which contains only 1,126 examples), when training together with the larger KonIQ and SPAQ datasets, compared to training  on LIVEC alone. These results emphasize the power of using CLIP as a prior with minimal adaptation, as it allows us to leverage CLIP’s perceptual knowledge, which is likely well-aligned with the shared properties across different datasets. When comparing our multi-dataset training against previous models on these datasets, the improvement is even more significant (e.g., a relative improvement of 5.6\% over the previous SOTA on EmoROI-6). Our multi-dataset models achieve improved SOTA results across all datasets and tasks. 

\section{Interpretability and Visualizations}
\label{sec:visualizations}

We visualize our model's attention heads to gain deeper insight into its decision-making process. Specifically, we focus on analyzing attention across all layers to identify the heads most critical for prediction. To achieve this, we employ an automated attention masking approach to systematically determine the importance of individual heads. First, an image is passed through the model to obtain a baseline prediction. Then, the same image is processed again, but this time the CLS token's attention for the target head is replaced with a uniform, equal-weighted map. By comparing the resulting prediction with the baseline (i.e., before and after masking), we can determine whether the head's attention is significant for prediction. A significant drop in performance indicates that the head is important. Repeating this process over many images and averaging the prediction differences allows us to quantify the importance of each head.

For the most influential (automatically-detected) heads, we select images that exhibit the largest prediction differences.
A few such images are presented in \cref{fig:visulaization},
along with differences between attention heatmaps of PerceptCLIP and the original CLIP. This comparison highlights the shift in attention, revealing how our fine-tuned model reallocates focus to different image regions, revealing the features it considers relevant for perceptual property prediction.

\begin{table}[th!]
\resizebox{0.48\textwidth}{!}{%
\begin{tabular}{lcccccc}
    \toprule
    \multirow{3}{*}{\shortstack{\textbf{Training} \\ \textbf{Procedure}}}
    & \multicolumn{2}{c}{\textbf{IQA}}  
    & \textbf{VEA} 
    & \multicolumn{2}{c}{\textbf{Memorability}} \\
    & \multicolumn{2}{c}{KonIQ-10k}  
    & EmoSet-8 
    & \multicolumn{2}{c}{LaMem} \\
    \cmidrule(lr){2-3} \cmidrule(lr){4-4} \cmidrule(lr){5-6} 
    & {SRCC$\uparrow$} & {PLCC$\uparrow$} 
    & {ACC$\uparrow$} 
    & {SRCC$\uparrow$} & {MSE$\downarrow$} \\
    \midrule
        \textbf{Full Fine-tuning CLIP} 
    & {0.888} & {0.904} 
    & {0.715} 
    & {0.618} & {0.0466} \\
        \textbf{Frozen CLIP} 
    & {0.894} & {0.913} 
    & {0.845} 
    & {0.691} & {0.0081} \\

    \textbf{LoRA for CLIP} 
    & \textbf{0.945} & \textbf{0.954} 
    & \textbf{0.853} 
    & \textbf{0.741} & \textbf{0.0069} \\
    \bottomrule
\end{tabular}%
}
\vspace{-0.0cm}
\caption{\textbf{Impact of Training Strategies.} We compare three training strategies for the CLIP visual encoder—full fine-tuning, freezing, and LoRA adaptation. The LoRA-based approach (which is used in our framework) outperforms both full fine-tuning and freezing, striking the best balance between adaptation and preserving CLIP’s perceptual prior.}
\vspace{-0.1cm}
\label{tab:train_method}
\end{table}

\cref{fig:visulaization} displays visual examples for both emotion and memorability predictions. The attention maps show that the model attends to semantically relevant regions for each emotion. For instance, in the image classified as fear-inducing, the model focuses on blood, a strong visual cue for fear. For anger, it attends to fire, while for excitement, it highlights the girl with an open mouth and a 'like' hand gesture (see more examples in Supplementary \cref{fig:visulaization_app_1,fig:visulaization_app_2}). For memorability, the model mainly focuses on the most distinctive and recognizable objects in the image. For example, in an image containing multiple elements, the model attends to chimneys with smoke, which may contribute significantly to memorability. Altogether, these demonstrate that our fine-tuned model successfully learns to focus on perceptually meaningful image features.

\vspace{0.1cm}
\section{Ablation Study}
\label{sec:ablations}
\vspace{0.1cm}

In this section, we conduct an ablation study to analyze the impact of different design choices in our approach. 

\vspace{-0.1cm}
\paragraph{Training Strategy.}
We aim to investigate how to effectively leverage CLIP’s perceptual prior while adapting it to different tasks.
We compare three different training strategies: (1) full fine-tuning of CLIP’s vision encoder, (2) freezing CLIP and training only the MLP head, and (3) our chosen approach -- fine-tuning CLIP by using LoRA adaptation (of the attention heads). As shown in \Cref{tab:train_method},  full fine-tuning significantly degrades the performance. This is likely due to catastrophic forgetting, where the model loses its strong pre-trained prior and fails to generalize effectively.
While training only the MLP (with frozen CLIP weights) yields better results than fully fine-tuning CLIP, it still falls short compared to using LoRA. Using
LoRA adaptation provides the best balance, allowing the model to fine-tune efficiently while preserving CLIP’s strong perceptual prior. Ablation on our two-step multi-dataset training can be found in the Supplementary \ref{sup:ablations}.

\begin{table}[th!]
\resizebox{0.48\textwidth}{!}{%
\begin{tabular}{lcccccc}
    \toprule
    \multirow{2}{*}{\textbf{Method}} 
    & \multicolumn{2}{c}{\textbf{IQA}}  
    & \textbf{VEA} 
    & \multicolumn{2}{c}{\textbf{Memorability}} \\
    & \multicolumn{2}{c}{KonIQ-10k}  
    & EmoSet-8 
    & \multicolumn{2}{c}{LaMem} \\
    \cmidrule(lr){2-3} \cmidrule(lr){4-4} \cmidrule(lr){5-6} 
    & {SRCC$\uparrow$} & {PLCC$\uparrow$} 
    & {ACC$\uparrow$} 
    & {SRCC$\uparrow$} & {MSE$\downarrow$} \\
    \midrule
    \textbf{Linear Layer} 
    & {0.943} & {0.953} 
    & {0.8526} 
    & 0.734 & 0.0077 \\
    \textbf{MLP (Ours)} 
    & \textbf{0.945} & \textbf{0.954} 
    & \textbf{0.8528} 
    & \textbf{0.741} & \textbf{0.0069} \\
    \bottomrule
\end{tabular}%
}
\vspace{0.05cm}
\caption{\textbf{Effect of Different Heads on Performance.} We compared the performance of a simple linear layer and the MLP head used in our main model. The choice of head has a minor effect on performance in the IQA and VEA tasks. However, the MLP head shows better performance on the memorability task.}
\label{tab:heads}

\end{table}

\vspace{-0.1cm}
\paragraph{MLP vs. Linear Layer}
We explored the impact of different head architectures by comparing a simple linear layer with the MLP head used in our main model. As shown in Table \ref{tab:heads}, the choice of head has a minor effect on performance in the IQA and VEA tasks, with both architectures performing competitively. However, the MLP head demonstrates better performance on the memorability task. In all tasks, both heads consistently achieve SOTA results using our approach.

\vspace{-0.15cm}
\paragraph{Backbone Choice.}
To assess the role of different vision backbones as perceptual priors, we experimented with  3 different  pre-trained backbone models (MAE, DINOv2, and CLIP, all using ViT-L/14), while keeping the rest of the architecture and LoRA configuration the same. As shown in Supplementary Table \ref{tab:backbone}, while our  approach performs well also with DINOv2, CLIP achieves the best results across all three tasks. This provides further evidence that CLIP effectively captures perceptual properties, making it a strong prior for perceptual tasks.

\vspace{-0.15cm}
\paragraph{Effect of ViT Size.}
To assess the impact of vision transformer size on performance, we compared three CLIP vision encoder variants: ViT-B/16, ViT-B/32, and ViT-L/14. The results, summarized in Supplementary \Cref{tab:clip_sizes}, indicate that ViT-B/16 outperforms ViT-B/32, suggesting that a finer patch resolution contributes to better perceptual understanding. Additionally, ViT-L/14 achieves the best results across all tasks, reinforcing our choice to use it as our backbone.

\section{Conclusion}
\label{sec:conclusion}

We propose a simple unified framework for visual perceptual tasks, which leverages the implicit yet rich \emph{perceptual knowledge} of CLIP. 
The human-annotated image captions in CLIP's training data contain also human sentiments and emotions, which is what probably makes CLIP a particularly strong prior for \emph{human judgment}.
We apply minimal task-specific adaptations, aiming to balance preserving CLIP’s strong perceptual prior, while allowing necessary task-specific flexibility. Remarkably, this lightweight yet effective framework produces models that achieve state-of-the-art performance on 3 important perceptual tasks, as well as demonstrates impressive cross-dataset generalization capabilities. This suggests that complex, task-specific designs and knowledge are not necessarily required for modeling perceptual tasks.

\clearpage

\section{Acknowledgments}
This research was funded by the European Union (ERC grant No. 101142115).

{
    \small
    \bibliographystyle{ieeenat_fullname}
    \bibliography{main}
}

\clearpage  
\appendix   

\twocolumn[{%
    \centering
    \LARGE \textbf{Supplementary Material} \par 
    \vspace{1em}  % Space below title
}]

\setcounter{section}{0}  % Reset section counter
\renewcommand{\thesection}{A\arabic{section}}  % Sections as A1, A2, ...

\setcounter{figure}{0}  % Reset figure counter
\renewcommand{\thefigure}{S\arabic{figure}}  % Figures as S1, S2, ...

\setcounter{table}{0}  % Reset table counter (optional)
\renewcommand{\thetable}{ST\arabic{table}}  % Tables as ST1, ST2, ...

\section{Further Technical Details}

\label{sup:Technical}

\paragraph{Data Augmentations.}In our training we used a common data augmentation techniques which are detailed below.  
\textbf{Image Quality Assessment:}  
Following \cite{xu2024boosting}, during training, each image is augmented into 3 to 10 replicas, depending on the dataset. Each replica undergoes independent random horizontal and vertical flips (p = 0.5) and random cropping to 224×224.  
\textbf{Predicting Image Memorability:}  
Before feeding the images into the model, we resize the shortest side of each image to 224 pixels while maintaining the aspect ratio. A center crop of size 224×224 is then applied.  
\textbf{Visual Emotion Analysis:}  
We apply data augmentation inspired by \cite{yang2023emoset}, including random resized cropping to 224×224 and horizontal flipping (p = 0.5). During inference, images are resized so that their shorter side is 224 pixels while preserving the aspect ratio, followed by a center crop.

\paragraph{4-parameter logistic regression correction.}
\label{4_parameter_correction}For the Image quality assessment task, we report Pearson’s Linear Correlation Coefficient (PLCC) with a 4-parameter logistic regression correction, as proposed by~\cite{antkowiak2000final} and following \cite{xu2024boosting}. The 4-parameter logistic function is defined as:  
 
\begin{equation}
y' = \beta_2 + \frac{\beta_1 - \beta_2}{1 + e^{-\left( \frac{x - \beta_3}{|\beta_4|} \right)}}
\end{equation}
where \( y' \) is the transformed prediction, \( x \) is the original model output, and \( \beta_1 \), \( \beta_2 \), \( \beta_3 \), and \( \beta_4 \) are the four parameters.  
This function is fitted to the model predictions using non-linear regression, optimizing its parameters to best match the subjective scores. Once the predictions are adjusted using this transformation, PLCC is computed to measure the correlation between the corrected predictions and human ratings.

\section{Additional Results}
\label{sup:results}

\subsection{Image Memorability Generalization Results}
\cref{sup:MEM_gen} illustrates the generalization capability of our model compared to two other models on the memorability task. ResMem\cite{needell2022embracing} and ViTMem\cite{hagen2023image} were trained using a combined dataset of LaMem (58,741 images) \cite{khosla2015understanding} and MemCat (10,000 images) \cite{goetschalckx2019memcat}, while our model was trained exclusively on the LaMem dataset. All models were evaluated on the Things memorability dataset. Our model outperforms the others, demonstrating superior generalization capabilities.
\begin{table}[h!]

\centering
\begin{tabular}{lc}
    \toprule
    \multirow{2}{*}{\textbf{Method}} & \textbf{THINGS} \\
    \cmidrule(r){2-2} 

    & {SRCC} \\ 
    \midrule
    resMem & 0.22 \\ 
    ViTMem & 0.30 \\ 
    \midrule
    \textbf{PerceptCLIP (Ours)} & \textbf{0.34} \\ 
    \bottomrule
\end{tabular}

\caption{\textbf{Generalization Results on Things Dataset.}
}
\label{sup:MEM_gen}
\end{table}

\subsection{Visual Emotion Analysis - Additional Metrics}

In addition to the results presented in the paper, we provide additional evaluation metrics in \cref{tab:EMO_metrics_1,tab:EMO_metrics_2}, comparing our model with two leading existing models. These metrics include:
\emph{\textbf{Mean Average Precision (mAP):}} Measures the area under the precision-recall curve, assessing the model’s ability to rank positive examples higher than negatives across different thresholds.
\emph{\textbf{Macro F1 Score (m-F1):}} Computes the F1 score for each class independently and averages them, treating all classes equally regardless of dataset imbalance.
\emph{\textbf{Weighted F1 Score (w-F1):}} The harmonic mean of precision and recall, weighted by class support, ensuring that both high precision and recall contribute fairly across imbalanced datasets. 

By incorporating these diverse metrics, we ensure a more comprehensive assessment of model performance beyond simple accuracy. Our model consistently outperforms the others across all metrics, demonstrating its robustness and effectiveness in handling complex classification tasks.

\begin{table}[h]
\resizebox{0.48\textwidth}{!}{%
\centering
\begin{tabular}{lcccc}
    \toprule
    \multirow{2}{*}{\textbf{Method}} 
    & \multicolumn{4}{c}{\textbf{EmoSet-8}}  \\ 
    \cmidrule(r){2-5} 
    % \cmidrule(r){6-9}  
    & \textbf{Accuracy} & \textbf{mAP} & \textbf{m-F1} & \textbf{w-F1} \\ 
    \midrule
    SimEmotion  
    & 0.7906 & 0.7387 & 0.7514 & 0.7894  \\ 
    MVP  
    & 0.8192 & 0.8003 & 0.8059 & 0.8047  \\ 
    \midrule
    \textbf{PerceptCLIP (Ours)}  
    & \textbf{0.8528} & \textbf{0.9336} & \textbf{0.8592} & \textbf{0.8526}  \\ 
    \bottomrule
\end{tabular}}
\caption{\textbf{Additional Metrics EmoSet.}}
\label{tab:EMO_metrics_1}
\end{table}

\begin{table}[h]
\resizebox{0.48\textwidth}{!}{%

\centering
\begin{tabular}{lccccc}
    \toprule
    \multirow{2}{*}{\textbf{Method}}  
    & \multicolumn{4}{c}{\textbf{EmotionROI-6}} \\ 
    \cmidrule(r){2-5} 
    % \cmidrule(r){6-9}  
    & \textbf{Accuracy} & \textbf{mAP} & \textbf{m-F1} & \textbf{w-F1} \\ 
    \midrule
    SimEmotion  
    & 0.7054 & 0.7054 & 0.7082 & 0.7027  \\ 
    MVP  
    & 0.7189 & 0.7116 & 0.7194 & 0.7148  \\ 
    \midrule
    \textbf{PerceptCLIP (Ours)}   
    & \textbf{0.7485} & \textbf{0.8081} & \textbf{0.7432} & \textbf{0.7445}  \\ 
    \bottomrule
\end{tabular}}
\caption{\textbf{Additional Metrics EmotionROI.}}
\label{tab:EMO_metrics_2}
\end{table}

\begin{table*}[h!]
\centering
\resizebox{\textwidth}{!}{%
\begin{tabular}{lcccccc||cccccc}
    \toprule
    \multirow{2}{*}{\textbf{Method}} & \multicolumn{2}{c}{\textbf{LIVEC}} & \multicolumn{2}{c}{\textbf{KonIQ-10k}} & \multicolumn{2}{c}{\textbf{SPAQ}} & \multicolumn{2}{c}{\textbf{LIVE}} & \multicolumn{2}{c}{\textbf{TID2013}} & \multicolumn{2}{c}{\textbf{KADID-10k}} \\
    \cmidrule(r){2-3} \cmidrule(r){4-5} \cmidrule(r){6-7} \cmidrule(r){8-9} \cmidrule(r){10-11} \cmidrule(r){12-13}
    & {SRCC} & {PLCC} & {SRCC} & {PLCC} & {SRCC} & {PLCC} & {SRCC} & {PLCC} & {SRCC} & {PLCC} & {SRCC} & {PLCC} \\
    \midrule
    \textbf{PerceptCLIP} & {0.908} & {0.923} & \textbf{0.945} & \textbf{0.954} & \textbf{0.930} & {0.933} & \textbf{0.980} & {0.982} & {0.887} & {0.908} & {0.952} & \textbf{0.956} \\
    \textbf{PerceptCLIP \small Multi-Dataset}
& \textbf{0.922} & \textbf{0.933} & {0.944} & {0.953} & \textbf{0.930} & \textbf{0.934} & \textbf{0.980} & \textbf{0.983} & \textbf{0.900} & \textbf{0.915} & \textbf{0.953} & \textbf{0.956} \\
    \bottomrule
\end{tabular}}
\caption{\textbf{Multi-Dataset vs. Single-Dataset Results on IQA:} We compare models trained on individual datasets with those trained on multiple datasets. Multi-dataset models were trained on either authentic distortions (LIVEC, KonIQ-10k, SPAQ) or synthetic distortions (LIVE, TID2013, KADID-10k). Multi-Dataset models improves performance on smaller datasets while maintaining comparable results on larger ones.}
\label{tab:Multi_IQA}
\end{table*}

\begin{table*}[h!]
    \centering
    \begin{minipage}{0.5\textwidth}
        \centering
        \resizebox{\textwidth}{!}{%
        \begin{tabular}{lcccc}
            \toprule
            \multirow{2}{*}{\textbf{Method}} 
            & \multicolumn{2}{c}{\textbf{LaMem}} 
            & \multicolumn{2}{c}{\textbf{THINGS}} \\ 
            \cmidrule(r){2-3} 
            \cmidrule(r){4-5}  
            & {SRCC} & {MSE} 
            & {SRCC} & {MSE}  \\ 
            \midrule
            \textbf{PerceptCLIP}  
            & \textbf{0.741} & \textbf{0.0069}  
            & {0.452} & {0.0058}  \\
            \textbf{PerceptCLIP \small Multi-Dataset}
            & {0.740} & \textbf{0.0069}  
            & \textbf{0.454} & \textbf{0.0054}  \\
            \bottomrule
        \end{tabular}%
        }
        \caption{\textbf{Multi-Dataset vs. Single-Dataset Training for Memorability Prediction:} The multi-dataset approach improves performance on THINGS while maintaining comparable results on LaMem.}
        \label{tab:Multi_MEM}
    \end{minipage}
    \hfill
    \begin{minipage}{0.48\textwidth}
        \centering
        \resizebox{\textwidth}{!}{%
        \begin{tabular}{lcc}
            \toprule
            {\textbf{Method}} 
            & {\textbf{EmoSet-8}} 
            & {\textbf{EmotionROI-6}} \\   
            \midrule
            \textbf{PerceptCLIP}  
            & \textbf{0.8528} 
            & {0.7485}  \\ 
            \textbf{PerceptCLIP \small Multi-Dataset}
            & {0.8511}  
            & \textbf{0.7591}   \\ 
            \bottomrule
        \end{tabular}%
        }
        \caption{\textbf{Multi-Dataset vs. Single-Dataset Training for Emotion Recognition:} Training EmotionROI-6 together with EmoSet-8 improves performance on EmotionROI-6 while maintaining comparable results to single-dataset training on EmoSet-8.}
        \label{tab:Multi_EMO}
    \end{minipage}
\end{table*}

\subsection{Multi-Dataset}
We conducted joint training on multiple datasets for the same task, where the CLIP LoRA adaptation is shared, and each dataset has its own MLP.
We present the results of these models on all datasets in \cref{tab:Multi_IQA,tab:Multi_EMO,tab:Multi_MEM}. These results show a significant improvement on smaller datasets, leveraging multi-dataset training with our approach, while achieving comparable results on larger datasets. Overall, our models trained on multiple datasets achieve state-of-the-art results across all tasks and datasets.

\section{Ablation Study}
\label{sup:ablations}
In this section, we present additional details from our ablation studies, including comprehensive tables, as well as further ablations of the multi-dataset training procedure.
\\\\
\noindent\textbf{{Different CLIP ViT Sizes.}} Table \ref{tab:clip_sizes} Shows the results of using different ViT sizes for the CLIP vision encoder backbone in our architecture. 

\noindent\textbf{{Different Vision Backbones.}} Table \ref{tab:backbone} shows the effect of different vision backbones on performance.

\paragraph{Multi-Dataset Training}
In our multi-dataset setting, we use two-step training procedure, where we first train the shared CLIP LoRA weights and the MLP for each dataset together with all datasets. In the second step, each dataset's MLP is trained separately while the CLIP LoRA weights are fixed and not optimized.
We have conducted an ablation study in~\cref{tab:multi_datasets_ablation} to demonstrate that the performance improvement is not solely due to the two-phase training procedure. We compare three approaches: (1) training a separate model for each dataset using our single-dataset approach, (2) training a separate model for each dataset using a two-phase procedure (first training the entire model, then fine-tuning only the MLP while keeping CLIP frozen), and (3) training a single model on multiple datasets using our multi-dataset approach with the two-phase training. The multi-dataset approach leads to better performance, while the two-phase training with a single dataset yields comparable results to the single-dataset approach.

\clearpage

\begin{table*}[h!]
\centering
\resizebox{0.65\textwidth}{!}{%
\begin{tabular}{lcccccc}
    \toprule
    \multirow{2}{*}{\textbf{Method}} 
    & \multicolumn{2}{c}{\textbf{IQA}}  
    & \textbf{VEA} 
    & \multicolumn{2}{c}{\textbf{Memorability}} \\
    & \multicolumn{2}{c}{KonIQ-10k}  
    & EmoSet-8 
    & \multicolumn{2}{c}{LaMem} \\
    \cmidrule(lr){2-3} \cmidrule(lr){4-4} \cmidrule(lr){5-6} 
    & {SRCC$\uparrow$} & {PLCC$\uparrow$} 
    & {ACC$\uparrow$} 
    & {SRCC$\uparrow$} & {MSE$\downarrow$} \\
    \midrule
    \textbf{ViT-B/32} 
    & 0.932 & 0.944 
    & 0.8303 
    & 0.724 & 0.0073 \\
    \textbf{ViT-B/16} 
    & {0.938} & {0.949} 
    & {0.8422} 
    & 0.731 & 0.0071 \\
    \textbf{ViT-L/14} 
    & \textbf{0.945} & \textbf{0.954} 
    & \textbf{0.8528} 
    & \textbf{0.741} & \textbf{0.0069} \\
    \bottomrule
\end{tabular}%
}

\parbox{0.65\textwidth}{
\caption{\textbf{Performance of Different CLIP Variants:} We trained our model using different sizes of ViT as the CLIP vision encoder. The results show that ViT-B/16 outperforms ViT-B/32, suggesting that a finer patch resolution improves perceptual understanding. ViT-L/14 achieves the best overall performance, reinforcing its use as our backbone model.}\label{tab:clip_sizes}}
\end{table*}

\begin{table*}[h!]
\centering
\resizebox{0.65\textwidth}{!}{%
\begin{tabular}{lcccccc}
    \toprule
    \multirow{3}{*}{\textbf{Backbone}} 
    & \multicolumn{2}{c}{\textbf{IQA}}  
    & \textbf{VEA} 
    & \multicolumn{2}{c}{\textbf{Memorability}} \\
    & \multicolumn{2}{c}{KonIQ-10k}  
    & EmoSet-8 
    & \multicolumn{2}{c}{LaMem} \\
    \cmidrule(lr){2-3} \cmidrule(lr){4-4} \cmidrule(lr){5-6} 
    & {SRCC$\uparrow$} & {PLCC$\uparrow$} 
    & {ACC$\uparrow$} 
    & {SRCC$\uparrow$} & {MSE$\downarrow$} \\
    \midrule
        \textbf{MAE} 
    & {0.924} & {0.935} 
    & {0.7681} 
    & {0.699} & {0.0078} \\
    \textbf{DinoV2} 
    & {0.937} & {0.950} 
    & {0.8252} 
    & {0.729} & {0.0071} \\
        \textbf{CLIP} 
    & \textbf{0.945} & \textbf{0.954} 
    & \textbf{0.8528} 
    & \textbf{0.741} & \textbf{0.0069} \\

    \bottomrule
\end{tabular}%
}
\parbox{0.65\textwidth}{\caption{ \textbf{Effect of Vision Backbones on Performance:} We trained our model with different backbones, applying LoRA to the q, k, and v layers. CLIP outperforms MAE and DINOv2 across all tasks, demonstrating its stronger perceptual prior.}\label{tab:backbone}}

\end{table*}

\begin{table*}[h]
\centering 

\resizebox{1\textwidth}{!}{%
\begin{tabular}{lccccccc}
    \toprule
    \multirow{3}{*}{\textbf{Method}}
    & \multicolumn{4}{c}{\textbf{IQA}}
    & \textbf{VEA}
    & \multicolumn{2}{c}{\textbf{Memorability prediction}}
    \\
        \cmidrule(r){2-5} 
    \cmidrule(r){6-6}
        \cmidrule(r){7-8}

    & \multicolumn{2}{c}{\textbf{LIVEC}}
    & \multicolumn{2}{c}{\textbf{TID2013}}
    & \textbf{EmoROI-6} 
    & \multicolumn{2}{c}{\textbf{THINGS}} \\ 
    \cmidrule(r){2-3} 
    \cmidrule(r){4-5} 
    \cmidrule(r){6-6}
    \cmidrule(r){7-8} 
    & {SRCC$\uparrow$} & {PLCC$\uparrow$}
    & {SRCC$\uparrow$} & {PLCC$\uparrow$} 
    & {ACC$\uparrow$}  
    & {SRCC$\uparrow$} & {MSE$\downarrow$}  \\ 
    \midrule
\textbf{PerceptCLIP \small Single-Dataset}
    & 0.908 & {0.923} 
    & 0.887 & 0.908
    & {0.7485}  
    & {0.452} & {0.0058} 
     \\
\textbf{PerceptCLIP \small 2-Phase Training Single-Dataset}
    & {0.907} & {0.923}
    & 0.889 & 0.909
    & {0.7500} 
    & {0.451} & {0.0055} 
    \\
\textbf{PerceptCLIP \small 2-Phase Training Multi-Dataset}
    & \textbf{0.922} & \textbf{0.933} 
    & \textbf{0.900} & \textbf{0.915} 

    & \textbf{0.7591} 
    & \textbf{0.454} &  \textbf{0.0054} 
    \\

    \bottomrule
\end{tabular}}
\caption{\textbf{Ablation Study for Multi-Dataset Training.} To demonstrate that the
performance improvement in our multi-dataset setting is not solely due to the two
phase training procedure, we compare three approaches:
(1) training a separate model for each dataset using our single-dataset approach, (2) training a separate model for each dataset using a two-phase procedure, and (3) training a single model on 
multiple datasets using our multi-dataset approach with the two-phase training. The multi-dataset approach leads 
to better performance, while the two-phase training with a single dataset yields comparable results to the single-dataset approach.}
\label{tab:multi_datasets_ablation}
\end{table*}

\clearpage
\begin{figure*}[h]
    \centering
    \includegraphics[width=1.\linewidth]{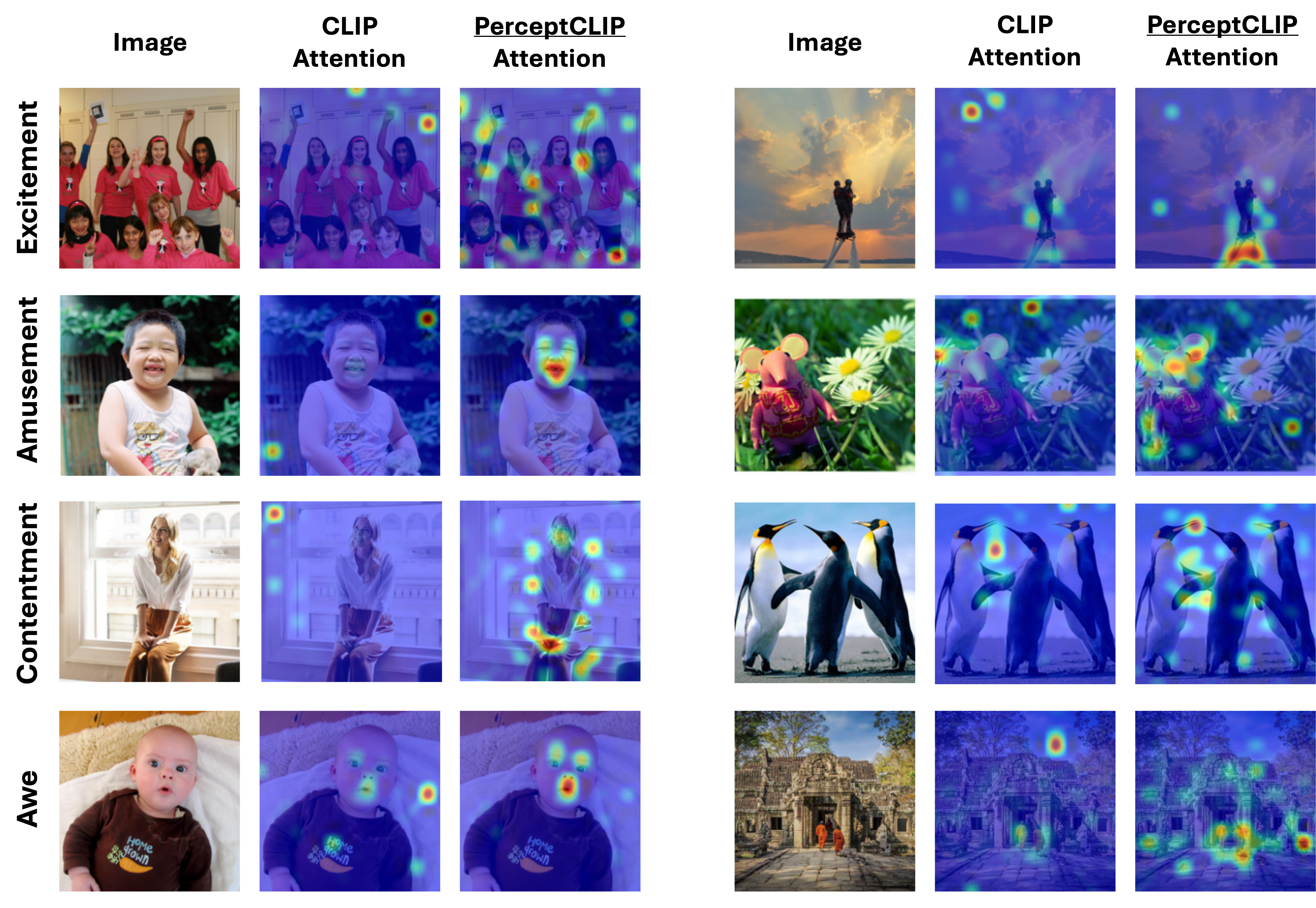}
\caption{\textbf{Attention Mask Visualization - Positive Emotions.} We present images alongside their corresponding attention maps from the pretrained CLIP vision encoder and our PerceptCLIP model (displaying results from critical attention heads that most influence the perceptual predictions). This highlights the shift in attention, revealing how our model reallocates focus to perceptually meaningful regions.}
    \label{fig:visulaization_app_1}
\end{figure*}

\begin{figure*}[h]
    \centering
    \includegraphics[width=1.\linewidth]{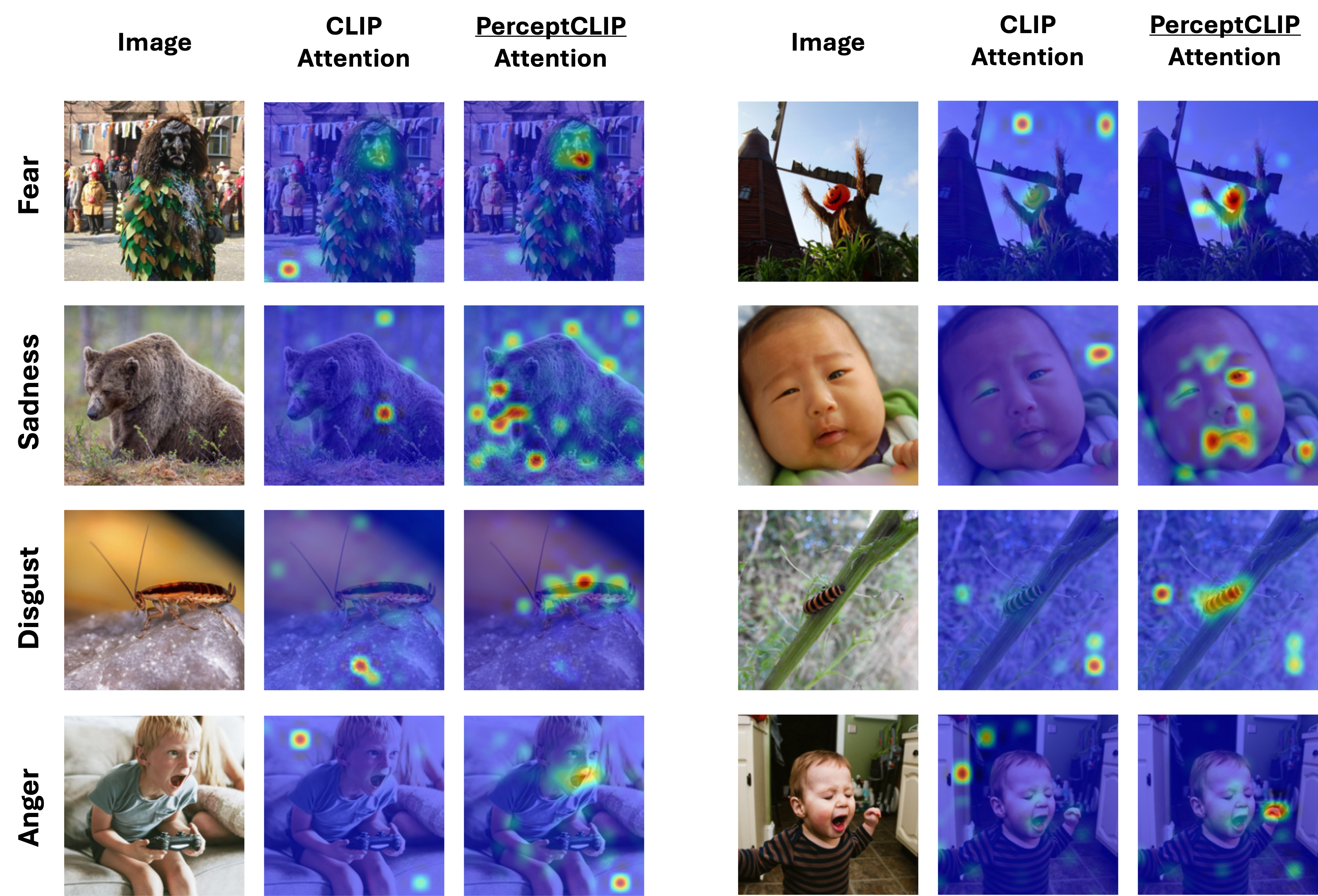}
\caption{\textbf{Attention Mask Visualization - Negative Emotions.} We present images alongside their corresponding attention maps from the pretrained CLIP vision encoder and our PerceptCLIP model (displaying results from critical attention heads that most influence the perceptual predictions). This highlights the shift in attention, revealing how our model reallocates focus to perceptually meaningful regions.}
    \label{fig:visulaization_app_2}
\end{figure*}

\end{document}